\documentclass[serv,nonblindrev]{informs-serv}

\OneAndAHalfSpacedXI

\usepackage{natbib}
\usepackage{booktabs}
\usepackage{amsmath}
\usepackage{amssymb}
\usepackage{graphicx}
\usepackage{float}
\usepackage{xcolor}
\definecolor{RepoLinkColor}{HTML}{7A2F2F}
\IfFileExists{orcidlink.sty}{\usepackage{orcidlink}}{}
\providecommand{\href}[2]{#2}
\makeatletter
\@ifundefined{orcidlink}
  {\newcommand{\orcidicon}[1]{\textsc{ORCID}\ }}
  {\newcommand{\orcidicon}[1]{\orcidlink{#1}\ }}
\newcommand{\orcidurl}[1]{\href{https://orcid.org/#1}{https://orcid.org/#1}}
\makeatother

\newcommand{\plotplaceholder}[4]{%
\begin{figure}[H]
\centering
\IfFileExists{#1}{%
\includegraphics[width=0.95\linewidth]{#1}%
}{%
\fbox{\parbox[c][0.32\textheight][c]{0.92\linewidth}{\centering
\textbf{Placeholder for #2}\\[2pt]
{\ttfamily\detokenize{#1}}}}%
}
\caption{#3}
\label{#4}
\end{figure}
}
\newcommand{\PaperKeywords}{stochastic optimizer selection, LR scaling, optimizer pool, dynamic optimizer switching, convergence acceleration, deep neural network optimization}

\bibpunct[, ]{(}{)}{,}{a}{}{,}%
\def\bibfont{\small}%

\TheoremsNumberedThrough
\ECRepeatTheorems
\EquationsNumberedThrough

\MANUSCRIPTNO{}
\renewcommand{\theARTICLETOP}{}
\def\theLRHFirstLine{Stamatis Mastromichalakis}
\def\theLRHSecondLine{}
\def\theRRHFirstLine{Stamatis Mastromichalakis}
\def\theRRHSecondLine{}

\begin{document}

\RUNAUTHOR{Stamatis Mastromichalakis}
\RUNTITLE{OptiRoulette}
\TITLE{OptiRoulette Optimizer: A New Stochastic Meta-Optimizer for up to 5.3x Faster Convergence}

\ARTICLEAUTHORS{%
\AUTHOR{Stamatis Mastromichalakis}
\AFF{Independent Researcher, tmnetworks.gr\\
\EMAIL{stamatis@tmnetwork.gr}\\
\orcidicon{0000-0003-1713-5078}\orcidurl{0000-0003-1713-5078}}
}

\ABSTRACT{%
This paper presents OptiRoulette, a stochastic meta-optimizer that selects update rules during training instead of fixing a single optimizer. The method combines warmup optimizer locking, random sampling from an active optimizer pool, compatibility-aware learning-rate scaling during optimizer transitions, and failure-aware pool replacement. OptiRoulette is implemented as a drop-in, \texttt{torch.optim.Optimizer}-compatible component and packaged for pip installation. We report completed 10-seed results on five image-classification suites: CIFAR-100, CIFAR-100-C, SVHN, Tiny ImageNet, and Caltech-256. Against a single-optimizer AdamW baseline, OptiRoulette improves mean test accuracy from 0.6734 to 0.7656 on CIFAR-100 (+9.22 percentage points), 0.2904 to 0.3355 on CIFAR-100-C (+4.52), 0.9667 to 0.9756 on SVHN (+0.89), 0.5669 to 0.6642 on Tiny ImageNet (+9.73), and 0.5946 to 0.6920 on Caltech-256 (+9.74). Its main advantage is convergence reliability at higher targets: it reaches CIFAR-100/CIFAR-100-C 0.75, SVHN 0.96, Tiny ImageNet 0.65, and Caltech-256 0.62 validation accuracy in 10/10 runs, while the AdamW baseline reaches none of these targets within budget. On shared targets, OptiRoulette also reduces time-to-target (e.g., Caltech-256 at 0.59: 25.7 vs 77.0 epochs). Paired-seed deltas are positive on all datasets; CIFAR-100-C test ROC-AUC is the only metric not statistically significant in the current 10-seed study.
}%

\maketitle
{\small
\vspace*{-1.2em}
\noindent\textbf{\textit{Key words:}} \textit{\PaperKeywords}\\[0.35em]
\noindent\textbf{Disclaimer:} The OptiRoulette name refers exclusively to a machine-learning optimizer and has no affiliation, sponsorship, or technical relation to roulette manufacturers, casinos, or any physical/software gambling products or services.\\[0.35em]
\clearpage
\noindent\textit{Author: Stamatis Mastromichalakis}\\
\textit{Official Repository: \href{https://github.com/MStamatis/OptiRoulette}{\textcolor{RepoLinkColor}{GitHub Repository}}}
}

\section{Introduction}
Optimizer choice remains a first-order determinant of deep-network training efficiency and final quality. In current practice, however, training typically relies on a single fixed optimizer (e.g., SGD-family or Adam-family) across all epochs, despite well-known stage-dependent behavior: adaptive methods often provide strong early progress, while non-adaptive methods may provide stronger late-stage generalization. This stage mismatch motivates optimizer policies that can evolve during training rather than remain static.

Prior work has explored several directions toward this goal. One line uses one-way transitions from adaptive to non-adaptive updates, such as SWATS \citealt{swats} and bounded-adaptive schemes such as AdaBound \citealt{adabound}. Another line combines update dynamics through wrappers like Lookahead \citealt{lookahead}. A more recent line explicitly models multi-optimizer interleaving and switching; for example, IOMT (\citealt{iomt}) and DOIT (\citealt{doit}) use surrogate performance models and acquisition-style selection rules to choose optimizers over training stages. In parallel, meta-learning approaches treat optimizer design itself as a learned problem (e.g., LSTM-based optimizers in \citealt{l2lopt}).

These advances are important, but they also expose a practical tradeoff: richer optimizer-selection policies can improve adaptation, yet they may increase system complexity, introduce additional tuning surfaces, and reduce plug-and-play usability in standard training pipelines. OptiRoulette is designed around a complementary point in this design space: a lightweight stochastic interleaving policy that remains optimizer-native and deployment-oriented, while still exploiting stage-dependent optimizer diversity.

Specifically, OptiRoulette is proposed as a standalone optimizer that combines (i) warmup locking, (ii) stochastic epoch-level sampling from an active optimizer pool, (iii) compatibility-aware learning-rate scaling during optimizer transitions, and (iv) failure-aware pool replacement. The optimizer is implemented as a \texttt{torch.optim.Optimizer}-compatible component intended for drop-in usage and packaging.

This paper makes four contributions. First, it formalizes the OptiRoulette optimization process as a stochastic optimizer-selection mechanism over an evolving active set. Second, it provides a theory-grounded interpretation of why the warmup-plus-interleaving regime can accelerate convergence in practice. Third, it reports completed 10-seed evidence across five image-classification suites (CIFAR-100, CIFAR-100-C, SVHN, Tiny ImageNet, and Caltech-256), with emphasis on time-to-target convergence, which is the main competitive advantage observed in this study (Section~\ref{sec:convergence_adv}). Fourth, it positions the resulting milestone behavior against current literature and public-archive reporting conventions.

\section{OptiRoulette Optimizer Design}
\subsection{State and Components}
Let $\mathcal{O} = \{o_1,\dots,o_K\}$ be the set of configured optimizers. At each training epoch $e$, OptiRoulette maintains:
\begin{enumerate}
\item an active set $\mathcal{A}_e \subseteq \mathcal{O}$ (from the pool manager),
\item a blocked set $\mathcal{B}_e \subseteq \mathcal{O}$,
\item a current optimizer $o_e \in \mathcal{A}_e$.
\end{enumerate}

The optimizer state machine is split into phases:
\begin{enumerate}
\item \textbf{Warmup phase}: optimizer is locked to a predefined optimizer (SGD in our experiments),
\item \textbf{Roulette phase}: optimizer is sampled randomly from candidates.
\end{enumerate}

\subsection{Random Selection Rule}
Candidate optimizers at epoch $e$ are:
\begin{equation}
\mathcal{C}_e = \mathcal{A}_e \setminus \mathcal{B}_e,
\end{equation}
with fallback to $\mathcal{O}\setminus \mathcal{B}_e$ if $\mathcal{C}_e$ is empty.

With \texttt{avoid\_repeat=true}, the previous optimizer is removed from candidates when possible:
\begin{equation}
\mathcal{C}'_e =
\begin{cases}
\mathcal{C}_e \setminus \{o_{e-1}\}, & \text{if } |\mathcal{C}_e|>1,\\
\mathcal{C}_e, & \text{otherwise}.
\end{cases}
\end{equation}

The selected optimizer is sampled uniformly:
\begin{equation}
o_e \sim \mathrm{Unif}(\mathcal{C}'_e).
\end{equation}

In the reported experiments, switching granularity is \texttt{epoch}; therefore one optimizer is selected per epoch (except during warmup lock), and all batches in that epoch use the same optimizer.

\subsection{Warmup and Phase Transition}
Warmup can terminate in two ways:
\begin{enumerate}
\item fixed epoch cutoff (\texttt{warmup\_epochs}),
\item plateau criterion on validation accuracy slope.
\end{enumerate}

In the completed runs, fixed warmup is used with \texttt{warmup\_epochs=17}, \texttt{warmup\_optimizer=sgd}, and \texttt{dropafter\_warmup=true}. Hence SGD is forced during early training and then excluded from roulette candidates after warmup.

\subsection{Reward and Pool Replacement}
After each epoch, the evaluation loop computes a reward for the selected optimizer:
\begin{equation}
r_e = 0.7\cdot(\mathrm{valAcc}_e - \mathrm{valAcc}^{\text{last}}_{o_e}) + 0.3\cdot(\mathrm{valAcc}_e - \mathrm{valAcc}^{\text{global-last}}),
\end{equation}
then clips $r_e \in [-1,1]$.

Pool replacement is triggered by failure conditions:
\begin{enumerate}
\item consecutive low rewards below a threshold,
\item catastrophic validation drop relative to best observed validation accuracy.
\end{enumerate}

The replacement active optimizer is selected from backup candidates, optionally using compatibility constraints and preferred swap rules. Learning rates can be adjusted across swaps via compatibility-aware scaling rules.

\subsection{Why OptiRoulette Can Converge Faster}
The observed acceleration can be interpreted as a stage-wise stochastic preconditioning effect.

Let $g_t=\nabla_\theta \mathcal{L}(\theta_t)$ and let optimizer $o_t$ apply update map $\phi_{o_t}$ with internal state $s_t^{(o_t)}$ and base learning rate $\eta_{o_t}$. The OptiRoulette update can be written as
\begin{equation}
\theta_{t+1}=\theta_t-\eta_{o_t}\,\phi_{o_t}(g_t,s_t^{(o_t)}).
\end{equation}
Conditioned on selection probabilities $\pi_t(i)=\Pr[o_t=i]$ over active candidates, the expected update is
\begin{equation}
\mathbb{E}[\Delta\theta_t\mid g_t]=-\sum_{i\in\mathcal{A}_t}\pi_t(i)\,\eta_i\,\phi_i(g_t,s_t^{(i)}),
\end{equation}
which behaves as a mixture of optimizer-specific descent geometries rather than a single fixed preconditioner.

In the reported configuration, this mixture is preceded by an explicit \textbf{17-epoch warmup} (\texttt{warmup\_epochs=17}) with \texttt{warmup\_optimizer=sgd} at LR $0.1$, then SGD is removed (\texttt{dropafter\_warmup=true}). This creates a two-stage dynamic:
\begin{enumerate}
\item \textbf{Fast basin entry}: high-step SGD warmup rapidly moves from random initialization to a useful attraction region.
\item \textbf{Refinement regime}: after warmup, selection is over \{Nadam, Adam, AdamW, Ranger, Adan, Lion\} with substantially smaller base LRs (Nadam/Adam/AdamW: $10^{-3}$, Ranger: $7\cdot10^{-4}$, Adan: $1.2\cdot10^{-4}$, Lion: $10^{-5}$), reducing oscillation while preserving optimizer diversity.
\end{enumerate}
Relative to warmup LR $0.1$, this induces an immediate post-warmup step-size contraction of approximately $10^2$--$10^4$, which helps convert rapid early progress into stable late-stage refinement.

Compatibility-aware LR scaling further regularizes transitions:
\begin{enumerate}
\item high-to-low-family transition scale: $0.01$ (\texttt{from\_high\_to\_low}),
\item low-to-high-family transition scale: $10.0$ (\texttt{from\_low\_to\_high}),
\item special cases: AdaHessian LR override $0.15$, Lion LR cap $3\cdot10^{-4}$.
\end{enumerate}
These rules are designed to limit destructive step-size discontinuities across optimizer families and to keep switched dynamics inside a stable trust region.

Finally, pool replacement and recovery settings (\texttt{failure\_threshold=-0.15}, \texttt{consecutive\_failure\_limit=3}, \texttt{catastrophic\_drop=0.3}, grace period 5 epochs, momentum transfer scaling $0.5$) reduce persistence of underperforming optimizer states. The net effect is a biased random process toward stable, high-payoff update dynamics, which is consistent with the faster empirical time-to-target convergence reported in Table~\ref{tab:convergence}.

\subsection{Implementation Notes}
For the current settings, roulette is random over a restricted active set rather than random over all configured optimizers. A consequence of epoch-level switching is that per-epoch optimizer distribution becomes one-hot, making per-epoch strategy entropy (logged as \texttt{nash\_conv}/\texttt{strategy\_entropy}) collapse to zero.

\section{Experimental Setup}\label{sec:setup}
\subsection{Data and Architectures}
Before reporting results, we summarize the concrete CNN and data-pipeline settings used in the completed runs, because convergence behavior is sensitive to both architecture and augmentation choices.

\begin{enumerate}
\item \textbf{CIFAR-100}: the completed run uses a ResNet-110 classifier (100 classes, initial channels 64), batch size 128, and a $0.9/0.1$ train/validation split with seed 42. The augmentation stack combines random crop, random horizontal flip, cutout (size 8), and mild color jitter.
\item \textbf{CIFAR-100-C}: the completed run follows the same ResNet-110 training configuration family as CIFAR-100, while final test evaluation is performed on the corruption-shifted CIFAR-100-C distribution. This preserves a comparable training protocol and isolates distribution-shift effects at test time.
\item \textbf{SVHN}: the completed run uses ResNet-110 with initial channels 16 (10 classes), batch size 128, and a $0.9/0.1$ split. Augmentation is dataset-specific (padding-4 crop, AutoAugment-SVHN policy, cutout size 16) and deliberately excludes horizontal flipping.
\item \textbf{Tiny ImageNet}: the completed run uses the SAMix architecture configuration (200 classes, initial channels 64), batch size 128, and a $0.9/0.1$ train/validation split with seed 42.
\item \textbf{Caltech-256}: the completed run uses ResNet-50 (257 classes), batch size 128, an initial $0.8$ train split, and then a $0.9/0.1$ train/validation split. The image pipeline uses size 64, random resized crop, random horizontal flip, color jitter, and cutout.
\end{enumerate}

\subsection{Optimization Configuration}
Common key settings across the five experiments:
\begin{enumerate}
\item \texttt{max\_epochs=100} (CIFAR-100, CIFAR-100-C, Caltech-256) and \texttt{max\_epochs=110} (SVHN, Tiny ImageNet),
\item \texttt{warmup\_epochs=17}, \texttt{warmup\_optimizer=sgd},
\item \texttt{switch\_granularity=epoch}, \texttt{switch\_every\_steps=1},
\item \texttt{switch\_probability=1.0}, \texttt{avoid\_repeat=true},
\item optimizer pool with 7 active optimizers: \{sgd, nadam, adam, adamw, ranger, adan, lion\}.
\end{enumerate}

In the completed runs, optimizer-side controls are intentionally aligned across datasets: \texttt{dropafter\_warmup=true}, \texttt{use\_scheduler=false}, and gradient clipping at 2.0. Representative base learning rates in the post-warmup pool are $10^{-3}$ (Adam/AdamW/Nadam), $7\cdot10^{-4}$ (Ranger), $1.2\cdot10^{-4}$ (Adan), and $10^{-5}$ (Lion).

Taken together, this setting bundle (pool composition, learning-rate vector, warmup/switching controls, and stabilization rules) defines the default OptiRoulette experimental preset in this paper. This default was selected from internal cross-dataset tuning as the most consistently effective and stable choice on the majority of the studied datasets.

The baseline \texttt{simple} mode uses a single optimizer. In all completed runs reported here, that optimizer is AdamW (verified in per-run logs). Comparisons therefore quantify the benefit of OptiRoulette optimizer logic against a standard fixed optimizer.

\subsection{Evaluation Protocol}
Each experiment uses 10 distinct seeds. We report means, standard deviations, and 95\% confidence intervals from the completed run suites for CIFAR-100, CIFAR-100-C, SVHN, Tiny ImageNet, and Caltech-256, using a consistent post-processing and statistical-analysis protocol across datasets.

Runs are paired by seed between modes: for each dataset, the same run index uses the same seed in \texttt{roulette} and \texttt{simple}. The mapping is identical across CIFAR-100, CIFAR-100-C, SVHN, Tiny ImageNet, and Caltech-256:
\begin{center}
\begin{tabular}{@{}ll@{\hspace{2em}}ll@{}}
\texttt{run01} & 2746317213 & \texttt{run06} & 127978094 \\
\texttt{run02} & 1181241943 & \texttt{run07} & 939042955 \\
\texttt{run03} & 958682846  & \texttt{run08} & 2340505846 \\
\texttt{run04} & 3163119785 & \texttt{run09} & 946785248 \\
\texttt{run05} & 1812140441 & \texttt{run10} & 2530876844 \\
\end{tabular}
\end{center}

\section{Results (Current Completed Runs)}
\subsection{Convergence Speed: Main Competitive Advantage}\label{sec:convergence_adv}
Using the architecture and optimization settings detailed in Section~\ref{sec:setup}, the dominant practical advantage of OptiRoulette in the current experiments is convergence speed to useful validation-accuracy regimes. Table~\ref{tab:convergence} reports headline first-hit milestones in epochs and wall-clock seconds.

\begin{table}[t]
\TABLE
{Headline Convergence Milestones (First Hit of Validation Accuracy Target)\label{tab:convergence}}
{\begin{tabular}{@{}llcccc@{}}
\toprule
Dataset & Mode & Target & Reach Rate & First Epoch ($\downarrow$) & First Time (s) ($\downarrow$) \\
\midrule
CIFAR-100 & Roulette & 0.65 & 10/10 & 18.8 & 1124.1 \\
CIFAR-100 & Simple (AdamW) & 0.65 & 10/10 & 25.1 & 1508.7 \\
CIFAR-100 & Roulette & 0.75 & 10/10 & 30.4 & 1912.0 \\
CIFAR-100 & Simple (AdamW) & 0.75 & 0/10 & -- & -- \\
CIFAR-100-C & Roulette & 0.65 & 10/10 & 18.8 & 1122.8 \\
CIFAR-100-C & Simple (AdamW) & 0.65 & 10/10 & 25.7 & 1536.9 \\
CIFAR-100-C & Roulette & 0.75 & 10/10 & 30.8 & 1935.8 \\
CIFAR-100-C & Simple (AdamW) & 0.75 & 0/10 & -- & -- \\
SVHN & Roulette & 0.95 & 10/10 & 18.8 & 887.3 \\
SVHN & Simple (AdamW) & 0.95 & 10/10 & 28.9 & 1369.9 \\
SVHN & Roulette & 0.96 & 10/10 & 27.1 & 1370.2 \\
SVHN & Simple (AdamW) & 0.96 & 0/10 & -- & -- \\
Tiny ImageNet & Roulette & 0.60 & 10/10 & 23.7 & 4127.1 \\
Tiny ImageNet & Simple (AdamW) & 0.60 & 0/10 & -- & -- \\
Tiny ImageNet & Roulette & 0.65 & 10/10 & 44.1 & 7826.4 \\
Tiny ImageNet & Simple (AdamW) & 0.65 & 0/10 & -- & -- \\
Caltech-256 & Roulette & 0.59 & 10/10 & 25.7 & 1054.4 \\
Caltech-256 & Simple (AdamW) & 0.59 & 10/10 & 77.0 & 3138.4 \\
Caltech-256 & Roulette & 0.62 & 10/10 & 30.3 & 1250.6 \\
Caltech-256 & Simple (AdamW) & 0.62 & 0/10 & -- & -- \\
\bottomrule
\end{tabular}}
{}
\end{table}

For shared targets, OptiRoulette reaches targets earlier on CIFAR-100/CIFAR-100-C, SVHN, and Caltech-256: CIFAR-100 at 0.65 (18.8 vs 25.1 epochs), CIFAR-100-C at 0.65 (18.8 vs 25.7), SVHN at 0.95 (18.8 vs 28.9), and Caltech-256 at 0.59 (25.7 vs 77.0, nearly $3\times$ faster). For harder targets (CIFAR-100 0.75, CIFAR-100-C 0.75, SVHN 0.96, Tiny ImageNet 0.65, Caltech-256 0.62), the baseline does not reach the same regime within budget. Using a budget-capped lower-bound framing for non-attained targets, the implied speedup reaches up to 5.3x (CIFAR-100/CIFAR-100-C target 0.70: 18.8 epochs vs baseline not reached by epoch 100).

To make the fast-convergence picture explicit across all completed datasets, Table~\ref{tab:convergence_fine} reports additional OptiRoulette milestones.

\begin{table}[t]
\TABLE
{Fine-Grained OptiRoulette Convergence Milestones\label{tab:convergence_fine}}
{\begin{tabular}{@{}lcccc@{}}
\toprule
Dataset & Target & Reach Rate & First Epoch ($\downarrow$) & First Time (s) ($\downarrow$) \\
\midrule
CIFAR-100 & 0.70 & 10/10 & 18.8 & 1124.1 \\
CIFAR-100 & 0.72 & 10/10 & 21.1 & 1276.3 \\
CIFAR-100 & 0.73 & 10/10 & 23.4 & 1432.0 \\
CIFAR-100 & 0.74 & 10/10 & 26.3 & 1630.7 \\
CIFAR-100 & 0.75 & 10/10 & 30.4 & 1912.0 \\
CIFAR-100-C & 0.70 & 10/10 & 18.8 & 1122.8 \\
CIFAR-100-C & 0.72 & 10/10 & 22.0 & 1335.4 \\
CIFAR-100-C & 0.73 & 10/10 & 22.5 & 1370.2 \\
CIFAR-100-C & 0.74 & 10/10 & 25.7 & 1592.4 \\
CIFAR-100-C & 0.75 & 10/10 & 30.8 & 1935.8 \\
SVHN & 0.93 & 10/10 & 17.7 & 819.7 \\
SVHN & 0.94 & 10/10 & 18.7 & 880.7 \\
SVHN & 0.95 & 10/10 & 18.8 & 887.3 \\
SVHN & 0.96 & 10/10 & 27.1 & 1370.2 \\
Tiny ImageNet & 0.60 & 10/10 & 23.7 & 4127.1 \\
Tiny ImageNet & 0.62 & 10/10 & 27.6 & 4831.5 \\
Tiny ImageNet & 0.63 & 10/10 & 31.8 & 5591.2 \\
Tiny ImageNet & 0.64 & 10/10 & 35.3 & 6227.7 \\
Tiny ImageNet & 0.65 & 10/10 & 44.1 & 7826.4 \\
Caltech-256 & 0.60 & 10/10 & 27.3 & 1121.7 \\
Caltech-256 & 0.62 & 10/10 & 30.3 & 1250.6 \\
Caltech-256 & 0.64 & 10/10 & 36.9 & 1545.6 \\
Caltech-256 & 0.66 & 10/10 & 43.9 & 1851.6 \\
Caltech-256 & 0.68 & 10/10 & 63.7 & 2698.7 \\
\bottomrule
\end{tabular}}
{}
\end{table}

Hence, OptiRoulette reaches high validation regimes with full reach rate across all five datasets: 74\%/75\% on CIFAR-100 at epochs 26.3/30.4, 74\%/75\% on CIFAR-100-C at 25.7/30.8, 0.96 on SVHN at epoch 27.1, 0.65 on Tiny ImageNet at epoch 44.1, and 0.68 on Caltech-256 at epoch 63.7 (all 10/10 where reported). This convergence behavior, rather than only final-point accuracy, is the main competitive differentiator of OptiRoulette.

For camera-ready assembly, we reserve per-dataset figure slots for two visual diagnostics: (i) a combined train/validation accuracy-loss panel and (ii) mean ROC curves across seeds. The paths below are placeholders and can be replaced with exported notebook figures.

\plotplaceholder
{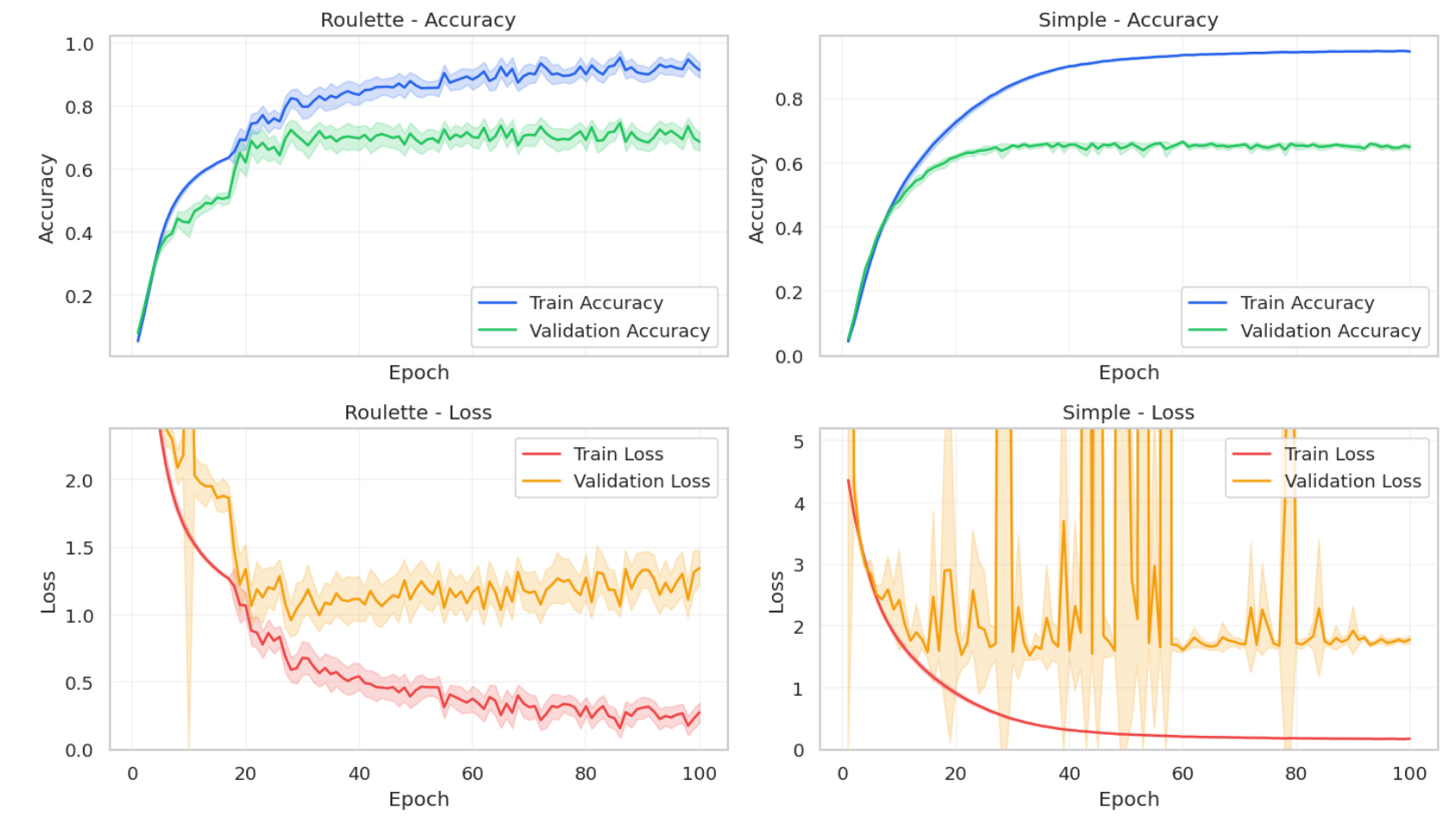}
{CIFAR-100 accuracy/loss panel}
{CIFAR-100: mean train/validation accuracy and loss trajectories across epochs for OptiRoulette and Simple (AdamW).}
{fig:cifar100_acc_loss_panel}

\plotplaceholder
{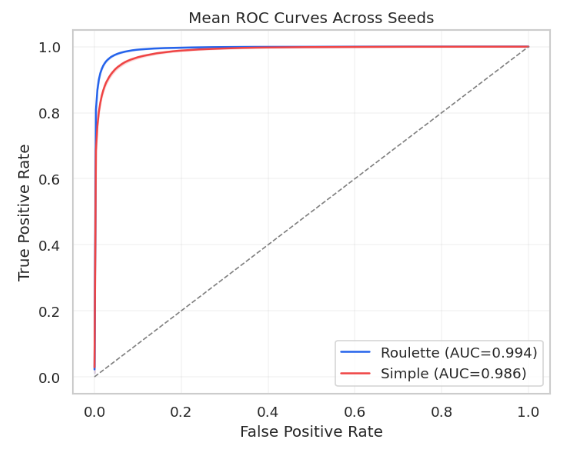}
{CIFAR-100 mean ROC}
{CIFAR-100: mean ROC curves across seeds for OptiRoulette and Simple (AdamW).}
{fig:cifar100_mean_roc}

\paragraph{CIFAR-100 visual analysis.}
Figure~\ref{fig:cifar100_acc_loss_panel} shows a clear separation in optimization dynamics after the early training phase: OptiRoulette reaches higher validation accuracy regimes earlier and then remains around a stable high plateau, whereas the fixed AdamW baseline saturates at a lower validation-accuracy level despite continued growth in training accuracy. The corresponding loss panels indicate that OptiRoulette drives validation loss to a lower and more stable operating range, while Simple (AdamW) exhibits substantially higher variance and frequent spikes in validation loss. Figure~\ref{fig:cifar100_mean_roc} is consistent with this behavior at evaluation time, with a higher mean ROC curve for OptiRoulette and a higher reported mean AUC (0.994 vs 0.986). Taken together, these plots support the interpretation that OptiRoulette's main gain on CIFAR-100 is faster and more reliable convergence to a stronger generalization regime.

\plotplaceholder
{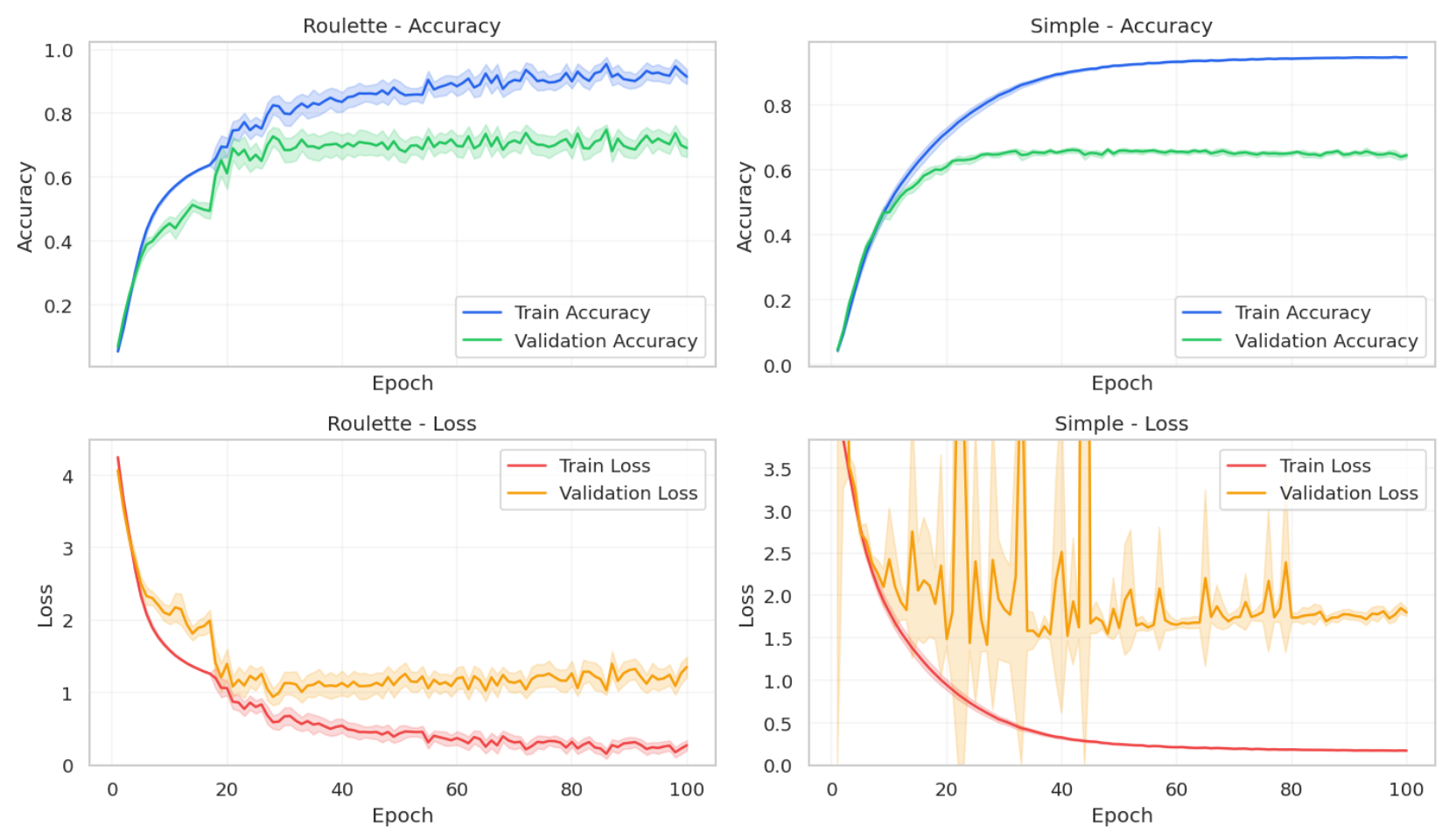}
{CIFAR-100-C accuracy/loss panel}
{CIFAR-100-C: mean train/validation accuracy and loss trajectories across epochs for OptiRoulette and Simple (AdamW).}
{fig:cifar100c_acc_loss_panel}

\plotplaceholder
{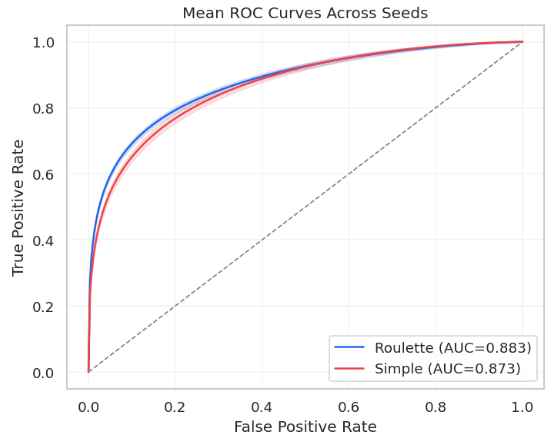}
{CIFAR-100-C mean ROC}
{CIFAR-100-C: mean ROC curves across seeds for OptiRoulette and Simple (AdamW).}
{fig:cifar100c_mean_roc}

\paragraph{CIFAR-100-C visual analysis.}
Figure~\ref{fig:cifar100c_acc_loss_panel} reproduces the same qualitative pattern under corruption shift: OptiRoulette reaches a higher validation-accuracy band earlier and remains consistently above the fixed-optimizer baseline across the training budget. In the loss panels, OptiRoulette shows a smoother and lower validation-loss trajectory, whereas Simple (AdamW) remains noisier with larger fluctuations and recurrent spikes, indicating weaker stability under shifted test conditions. Figure~\ref{fig:cifar100c_mean_roc} provides aligned ranking evidence at the classifier level, with OptiRoulette yielding a higher mean ROC curve and higher mean AUC (0.883 vs 0.873). Overall, the CIFAR-100-C plots support that the convergence and stability advantage of OptiRoulette persists beyond the in-distribution setting.

\plotplaceholder
{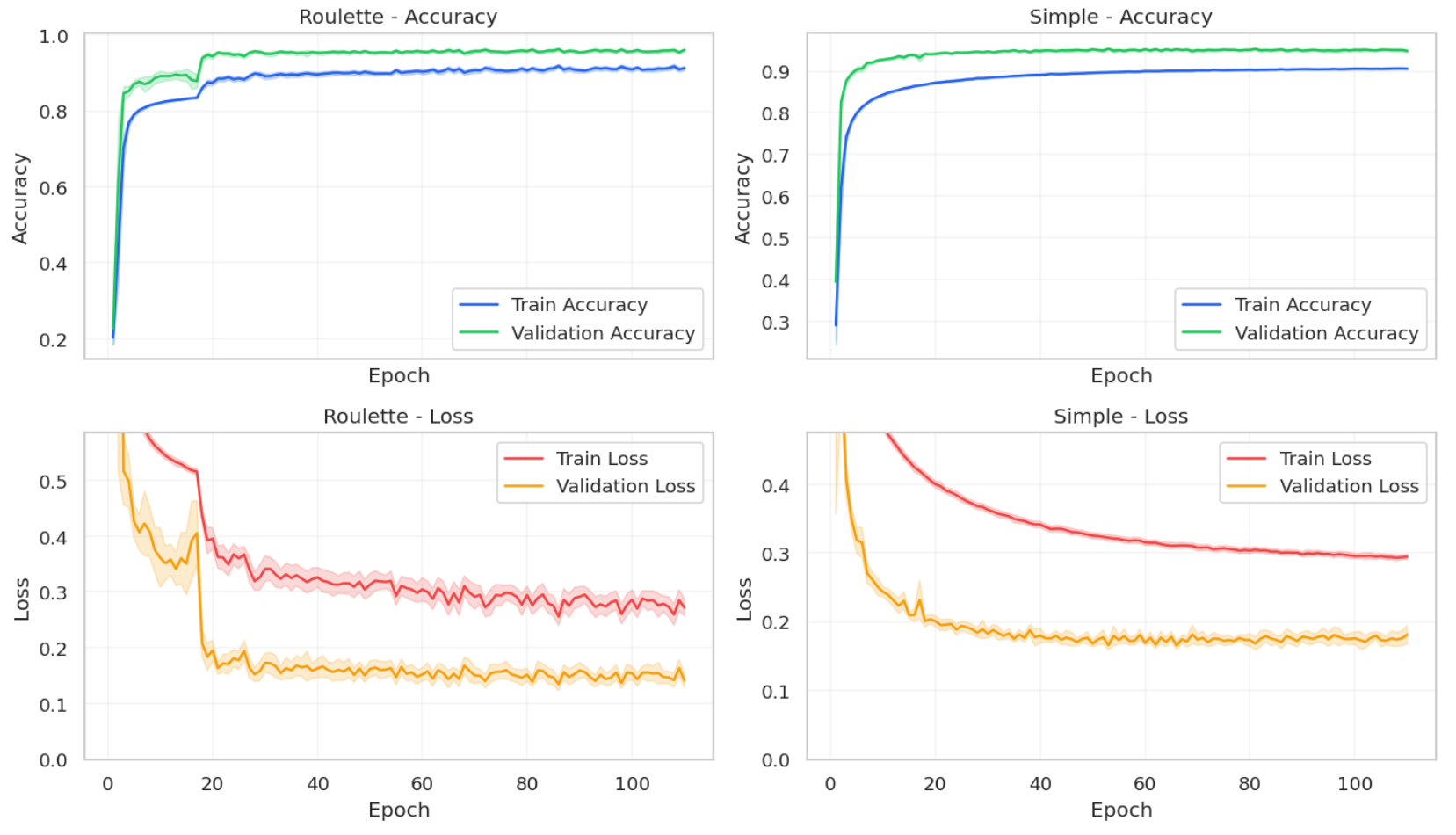}
{SVHN accuracy/loss panel}
{SVHN: mean train/validation accuracy and loss trajectories across epochs for OptiRoulette and Simple (AdamW).}
{fig:svhn_acc_loss_panel}

\plotplaceholder
{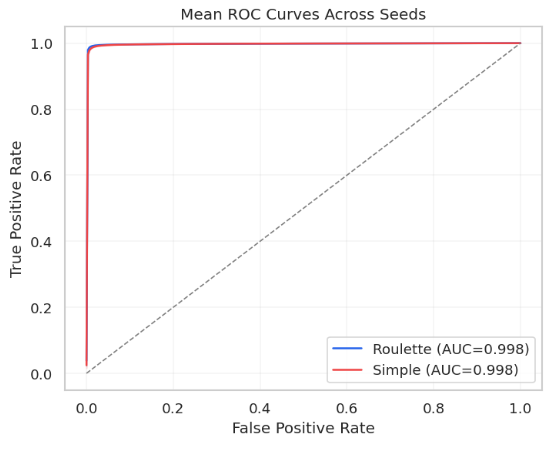}
{SVHN mean ROC}
{SVHN: mean ROC curves across seeds for OptiRoulette and Simple (AdamW).}
{fig:svhn_mean_roc}

\paragraph{SVHN visual analysis.}
Figure~\ref{fig:svhn_acc_loss_panel} indicates that both methods converge quickly on SVHN, but OptiRoulette still reaches high-validation regimes earlier and stabilizes at a slightly stronger operating point. A clear phase transition is visible around the warmup boundary, after which OptiRoulette's validation loss drops and remains low with limited variance. In contrast, Simple (AdamW) improves more gradually and plateaus at a marginally weaker level. Figure~\ref{fig:svhn_mean_roc} shows near-overlapping ROC curves and identical rounded mean AUC (0.998 vs 0.998), so the practical advantage on SVHN is primarily faster time-to-target and small but consistent accuracy gains rather than a large ROC-shape separation.

\plotplaceholder
{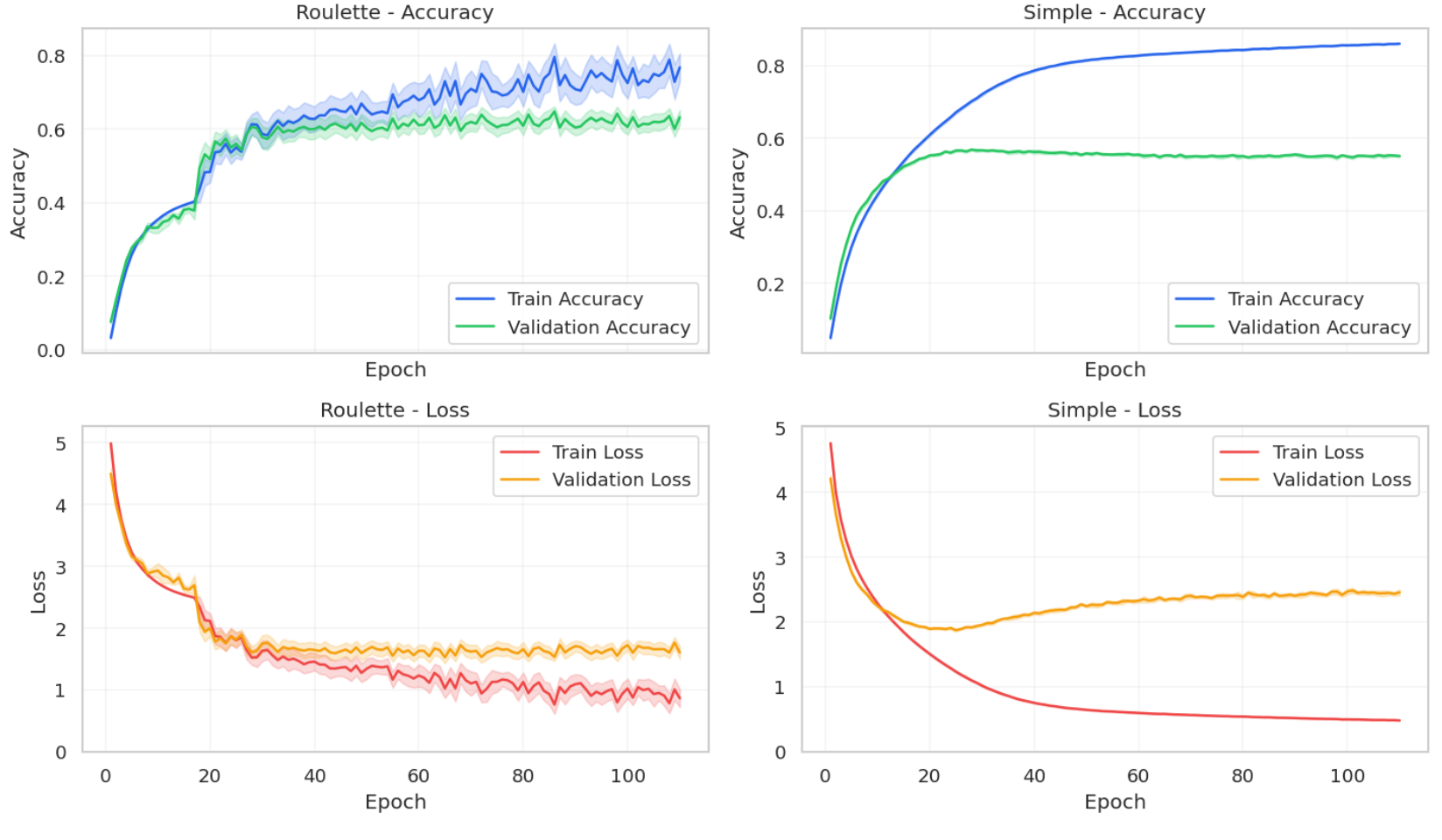}
{Tiny ImageNet accuracy/loss panel}
{Tiny ImageNet: mean train/validation accuracy and loss trajectories across epochs for OptiRoulette and Simple (AdamW).}
{fig:tinyimagenet_acc_loss_panel}

\plotplaceholder
{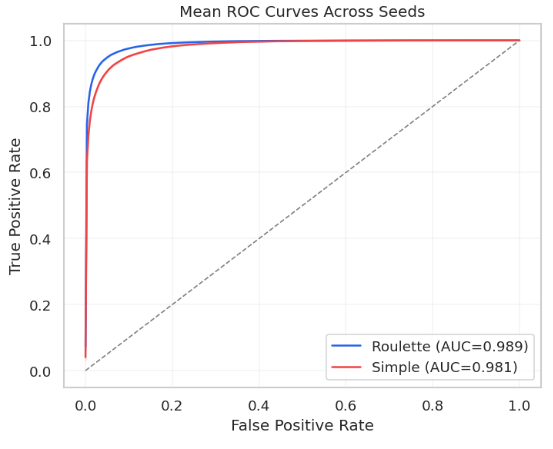}
{Tiny ImageNet mean ROC}
{Tiny ImageNet: mean ROC curves across seeds for OptiRoulette and Simple (AdamW).}
{fig:tinyimagenet_mean_roc}

\paragraph{Tiny ImageNet visual analysis.}
Figure~\ref{fig:tinyimagenet_acc_loss_panel} highlights a strong separation in generalization dynamics on the most challenging setting of the current study. OptiRoulette continues to lift validation accuracy after the early phase and reaches a clearly higher plateau, while Simple (AdamW) saturates earlier at a substantially lower validation-accuracy level. The loss plots reinforce this interpretation: in the baseline, validation loss declines initially but then trends upward and remains high, consistent with overfitting under a fixed optimizer policy; OptiRoulette maintains a lower and more controlled validation-loss trajectory throughout training. Figure~\ref{fig:tinyimagenet_mean_roc} is aligned with the same ranking, with a visibly better mean ROC curve and higher mean AUC for OptiRoulette (0.989 vs 0.981). These plots support the claim that OptiRoulette's convergence advantage is especially pronounced when optimization and generalization are both difficult.

\plotplaceholder
{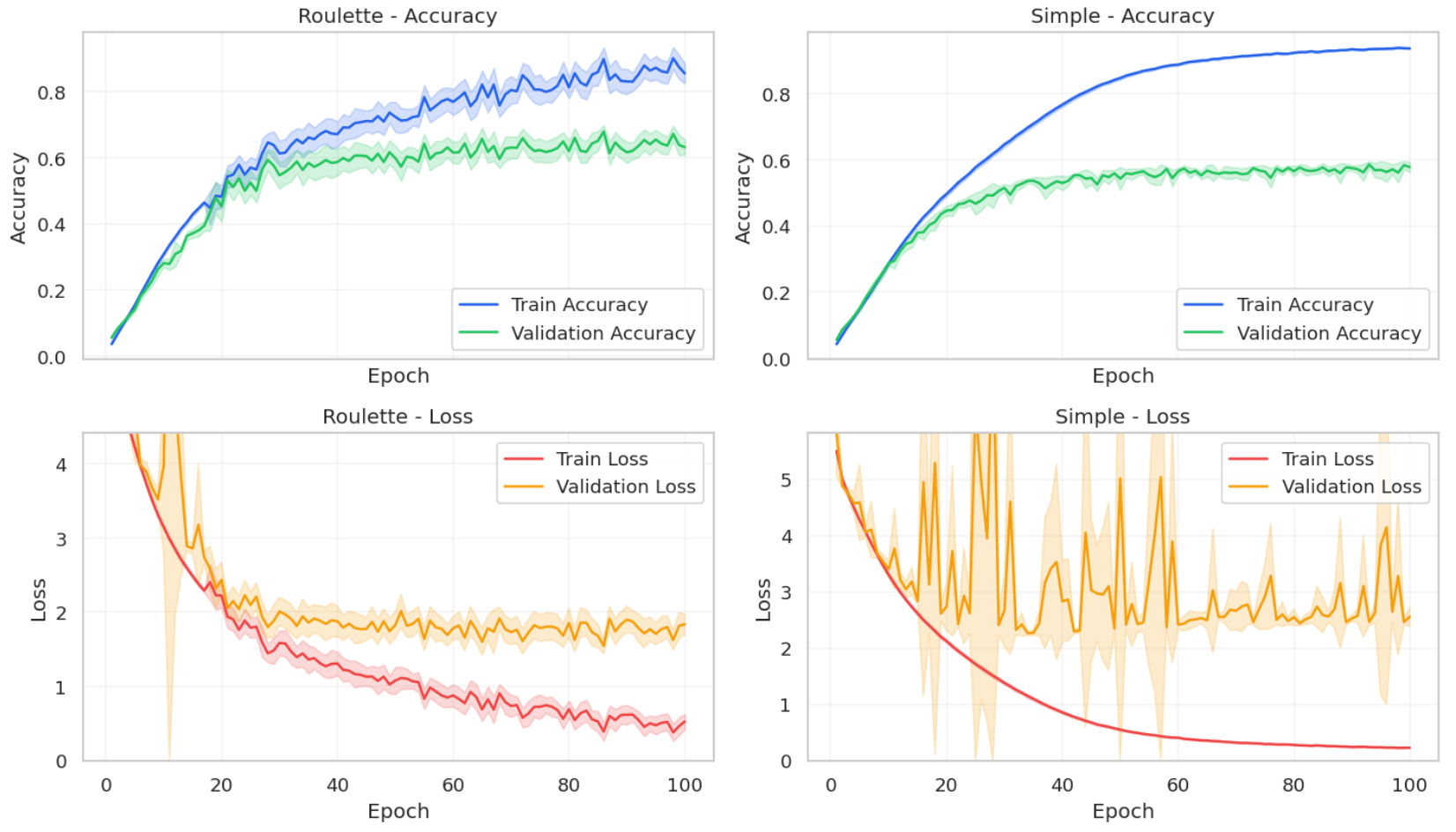}
{Caltech-256 accuracy/loss panel}
{Caltech-256: mean train/validation accuracy and loss trajectories across epochs for OptiRoulette and Simple (AdamW).}
{fig:caltech256_acc_loss_panel}

\plotplaceholder
{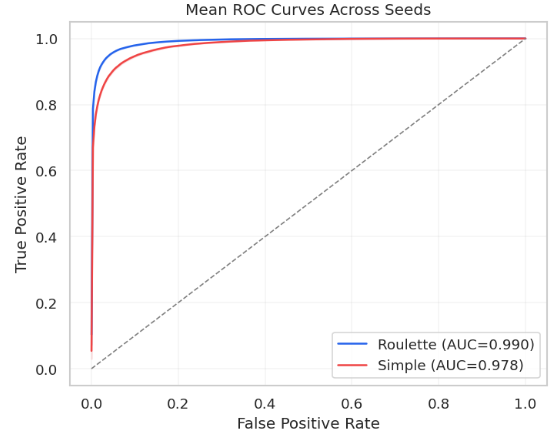}
{Caltech-256 mean ROC}
{Caltech-256: mean ROC curves across seeds for OptiRoulette and Simple (AdamW).}
{fig:caltech256_mean_roc}

\paragraph{Caltech-256 visual analysis.}
Figure~\ref{fig:caltech256_acc_loss_panel} shows a strong convergence-gap pattern in favor of OptiRoulette: validation accuracy rises to a higher regime substantially earlier and remains clearly above the fixed AdamW baseline throughout training. The loss trajectories provide complementary evidence. Under OptiRoulette, validation loss decreases to a lower and relatively stable band, while the Simple (AdamW) run exhibits larger volatility and persistent high-amplitude spikes, indicating less robust optimization dynamics. Figure~\ref{fig:caltech256_mean_roc} confirms the same ranking from the ROC perspective, with a higher mean ROC curve and higher mean AUC for OptiRoulette (0.990 vs 0.978). Together, these plots support the claim that the optimizer-switching mechanism yields both faster attainment of useful regimes and better final discrimination quality on Caltech-256.

\subsection{Aggregate Performance}
Table~\ref{tab:main_results} summarizes aggregate metrics from the five completed 10-seed studies.

\begin{table}[t]
\TABLE
{Main Results Across Completed 10-Seed Runs\label{tab:main_results}}
{\begin{tabular}{@{}llcccc@{}}
\toprule
Dataset & Mode & Val Acc ($\uparrow$) & Test Acc ($\uparrow$) & Test ROC-AUC ($\uparrow$) & Duration (s) \\
\midrule
CIFAR-100 & Roulette & $0.7705 \pm 0.0018$ & $0.7656 \pm 0.0024$ & $0.99396 \pm 0.00020$ & $6678.7 \pm 73.6$ \\
CIFAR-100 & Simple (AdamW) & $0.6792 \pm 0.0033$ & $0.6734 \pm 0.0038$ & $0.98640 \pm 0.00085$ & $5993.1 \pm 29.0$ \\
CIFAR-100-C & Roulette & $0.7722 \pm 0.0031$ & $0.3355 \pm 0.0046$ & $0.88291 \pm 0.00628$ & $6658.0 \pm 61.1$ \\
CIFAR-100-C & Simple (AdamW) & $0.6781 \pm 0.0033$ & $0.2904 \pm 0.0068$ & $0.87341 \pm 0.00779$ & $5963.5 \pm 25.6$ \\
SVHN & Roulette & $0.9666 \pm 0.0005$ & $0.9756 \pm 0.0004$ & $0.99792 \pm 0.00005$ & $6234.6 \pm 104.7$ \\
SVHN & Simple (AdamW) & $0.9578 \pm 0.0006$ & $0.9667 \pm 0.0013$ & $0.99769 \pm 0.00007$ & $5206.5 \pm 26.5$ \\
Tiny ImageNet & Roulette & $0.6675 \pm 0.0020$ & $0.6642 \pm 0.0020$ & $0.98854 \pm 0.00013$ & $19729.8 \pm 80.4$ \\
Tiny ImageNet & Simple (AdamW) & $0.5735 \pm 0.0010$ & $0.5669 \pm 0.0030$ & $0.98062 \pm 0.00053$ & $18993.1 \pm 26.0$ \\
Caltech-256 & Roulette & $0.7001 \pm 0.0020$ & $0.6920 \pm 0.0045$ & $0.98962 \pm 0.00026$ & $4251.7 \pm 127.1$ \\
Caltech-256 & Simple (AdamW) & $0.6014 \pm 0.0057$ & $0.5946 \pm 0.0053$ & $0.97786 \pm 0.00075$ & $4071.0 \pm 85.2$ \\
\bottomrule
\end{tabular}}
{\footnotesize Val/Test metrics are mean $\pm$ 95\% CI. Duration is mean $\pm$ std from run-level cumulative training-time logs.}
\end{table}

OptiRoulette improves mean test accuracy by $+0.0922$ on CIFAR-100 (+13.69\% relative), $+0.0452$ on CIFAR-100-C (+15.56\%), $+0.0089$ on SVHN (+0.92\%), $+0.0973$ on Tiny ImageNet (+17.16\%), and $+0.0974$ on Caltech-256 (+16.38\%). Runtime overhead is +11.01\% (CIFAR-100), +11.64\% (CIFAR-100-C), +19.75\% (SVHN), +3.88\% (Tiny ImageNet), and +4.44\% (Caltech-256).

\subsection{Additional Test Metrics: Precision, Recall, and F1}
Table~\ref{tab:main_results_prf} adds test precision/recall/F1 across all completed datasets. Because the current implementation uses weighted averaging for roulette and macro averaging for simple mode, these metrics should be interpreted as complementary descriptive evidence (see Threats to Validity), while accuracy and ROC-AUC remain the primary like-for-like comparison metrics.

\begin{table}[t]
\TABLE
{Additional Test Metrics Across Completed 10-Seed Runs\label{tab:main_results_prf}}
{\begin{tabular}{@{}llccc@{}}
\toprule
Dataset & Mode & Test Precision ($\uparrow$) & Test Recall ($\uparrow$) & Test F1 ($\uparrow$) \\
\midrule
CIFAR-100 & Roulette & $0.7685 \pm 0.0026$ & $0.7656 \pm 0.0024$ & $0.7657 \pm 0.0024$ \\
CIFAR-100 & Simple (AdamW) & $0.6985 \pm 0.0027$ & $0.6734 \pm 0.0038$ & $0.6732 \pm 0.0040$ \\
CIFAR-100-C & Roulette & $0.4509 \pm 0.0037$ & $0.3355 \pm 0.0046$ & $0.3572 \pm 0.0046$ \\
CIFAR-100-C & Simple (AdamW) & $0.4103 \pm 0.0053$ & $0.2904 \pm 0.0068$ & $0.3035 \pm 0.0067$ \\
SVHN & Roulette & $0.9757 \pm 0.0004$ & $0.9756 \pm 0.0004$ & $0.9756 \pm 0.0004$ \\
SVHN & Simple (AdamW) & $0.9648 \pm 0.0013$ & $0.9644 \pm 0.0016$ & $0.9645 \pm 0.0013$ \\
Tiny ImageNet & Roulette & $0.6684 \pm 0.0019$ & $0.6642 \pm 0.0020$ & $0.6636 \pm 0.0020$ \\
Tiny ImageNet & Simple (AdamW) & $0.5990 \pm 0.0021$ & $0.5669 \pm 0.0030$ & $0.5664 \pm 0.0027$ \\
Caltech-256 & Roulette & $0.7058 \pm 0.0040$ & $0.6920 \pm 0.0045$ & $0.6909 \pm 0.0045$ \\
Caltech-256 & Simple (AdamW) & $0.6128 \pm 0.0063$ & $0.5584 \pm 0.0049$ & $0.5583 \pm 0.0065$ \\
\bottomrule
\end{tabular}}
{\footnotesize Entries are mean $\pm$ 95\% CI.}
\end{table}

\begin{table}[t]
\TABLE
{Paired t-Tests for Test Precision, Recall, and F1 (Roulette -- Simple)\label{tab:paired_ttest_prf}}
{\begin{tabular}{@{}llccc@{}}
\toprule
Dataset & Metric & $\Delta$ Mean $\pm$ 95\% CI & $t(9)$ & $p$-value \\
\midrule
CIFAR-100 & Test Precision & $0.06998 \pm 0.00305$ & 45.00 & $6.61\times10^{-12}$ \\
CIFAR-100 & Test Recall & $0.09215 \pm 0.00348$ & 51.85 & $1.85\times10^{-12}$ \\
CIFAR-100 & Test F1 & $0.09253 \pm 0.00367$ & 49.40 & $2.86\times10^{-12}$ \\
CIFAR-100-C & Test Precision & $0.04067 \pm 0.00696$ & 11.45 & $1.15\times10^{-6}$ \\
CIFAR-100-C & Test Recall & $0.04518 \pm 0.00884$ & 10.02 & $3.52\times10^{-6}$ \\
CIFAR-100-C & Test F1 & $0.05367 \pm 0.00920$ & 11.44 & $1.16\times10^{-6}$ \\
SVHN & Test Precision & $0.01084 \pm 0.00127$ & 16.70 & $4.43\times10^{-8}$ \\
SVHN & Test Recall & $0.01127 \pm 0.00175$ & 12.64 & $4.95\times10^{-7}$ \\
SVHN & Test F1 & $0.01117 \pm 0.00136$ & 16.07 & $6.18\times10^{-8}$ \\
Tiny ImageNet & Test Precision & $0.06943 \pm 0.00210$ & 64.83 & $2.50\times10^{-13}$ \\
Tiny ImageNet & Test Recall & $0.09726 \pm 0.00249$ & 76.62 & $5.56\times10^{-14}$ \\
Tiny ImageNet & Test F1 & $0.09719 \pm 0.00231$ & 82.48 & $2.87\times10^{-14}$ \\
Caltech-256 & Test Precision & $0.09301 \pm 0.00618$ & 29.51 & $2.88\times10^{-10}$ \\
Caltech-256 & Test Recall & $0.13360 \pm 0.00659$ & 39.75 & $2.01\times10^{-11}$ \\
Caltech-256 & Test F1 & $0.13259 \pm 0.00672$ & 38.66 & $2.57\times10^{-11}$ \\
\bottomrule
\end{tabular}}
{}
\end{table}

\subsection{Paired-Seed Effect Size}
Because each seed is evaluated in both modes, paired differences provide a stronger comparison. Table~\ref{tab:paired} reports roulette minus simple deltas.

\begin{table}[t]
\TABLE
{Paired Seed Deltas (Roulette -- Simple)\label{tab:paired}}
{\begin{tabular}{@{}lccc@{}}
\toprule
Dataset & $\Delta$ Val Acc & $\Delta$ Test Acc & $\Delta$ Test ROC-AUC \\
\midrule
CIFAR-100 & $0.0913 \pm 0.0033$ & $0.0922 \pm 0.0035$ & $0.0076 \pm 0.0008$ \\
CIFAR-100-C & $0.0941 \pm 0.0055$ & $0.0452 \pm 0.0088$ & $0.0095 \pm 0.0097$ \\
SVHN & $0.0088 \pm 0.0010$ & $0.0089 \pm 0.0014$ & $0.00023 \pm 0.00005$ \\
Tiny ImageNet & $0.0940 \pm 0.0024$ & $0.0973 \pm 0.0025$ & $0.0079 \pm 0.0006$ \\
Caltech-256 & $0.0987 \pm 0.0058$ & $0.0974 \pm 0.0066$ & $0.0118 \pm 0.0009$ \\
\bottomrule
\end{tabular}}
{\footnotesize Entries are mean paired delta $\pm$ 95\% CI over 10 matched seeds.}
\end{table}

All confidence intervals remain strictly positive, indicating robust gains across seeds. For formal hypothesis testing, Table~\ref{tab:paired_ttest} reports paired two-sided t-tests on the same matched seeds.

\begin{table}[t]
\TABLE
{Paired t-Test Results (Roulette -- Simple, 10 Matched Seeds)\label{tab:paired_ttest}}
{\begin{tabular}{@{}llccc@{}}
\toprule
Dataset & Metric & $\Delta$ Mean $\pm$ 95\% CI & $t(9)$ & $p$-value \\
\midrule
CIFAR-100 & Val Acc & $0.0913 \pm 0.0033$ & 54.58 & $1.17\times10^{-12}$ \\
CIFAR-100 & Test Acc & $0.0922 \pm 0.0035$ & 51.85 & $1.85\times10^{-12}$ \\
CIFAR-100 & Test ROC-AUC & $0.00756 \pm 0.00082$ & 18.10 & $2.19\times10^{-8}$ \\
CIFAR-100-C & Val Acc & $0.0941 \pm 0.0055$ & 33.31 & $9.77\times10^{-11}$ \\
CIFAR-100-C & Test Acc & $0.04518 \pm 0.00884$ & 10.02 & $3.52\times10^{-6}$ \\
CIFAR-100-C & Test ROC-AUC & $0.00950 \pm 0.00970$ & 1.92 & $8.70\times10^{-2}$ \\
SVHN & Val Acc & $0.00880 \pm 0.00096$ & 17.99 & $2.30\times10^{-8}$ \\
SVHN & Test Acc & $0.00893 \pm 0.00138$ & 12.73 & $4.66\times10^{-7}$ \\
SVHN & Test ROC-AUC & $0.000233 \pm 0.000052$ & 8.83 & $9.96\times10^{-6}$ \\
Tiny ImageNet & Val Acc & $0.0940 \pm 0.0024$ & 75.85 & $6.09\times10^{-14}$ \\
Tiny ImageNet & Test Acc & $0.0973 \pm 0.0025$ & 76.62 & $5.56\times10^{-14}$ \\
Tiny ImageNet & Test ROC-AUC & $0.00792 \pm 0.00057$ & 27.13 & $6.08\times10^{-10}$ \\
Caltech-256 & Val Acc & $0.09869 \pm 0.00581$ & 33.30 & $9.79\times10^{-11}$ \\
Caltech-256 & Test Acc & $0.09740 \pm 0.00662$ & 28.82 & $3.55\times10^{-10}$ \\
Caltech-256 & Test ROC-AUC & $0.01176 \pm 0.00088$ & 26.23 & $8.20\times10^{-10}$ \\
\bottomrule
\end{tabular}}
{}
\end{table}

Most tested metric deltas are statistically significant ($p<0.001$). The one exception is CIFAR-100-C test ROC-AUC ($p=0.08697$), where the current 10-seed evidence is inconclusive.

\subsection{Validation Milestones}
Milestone analysis from per-epoch logs is consistent with the convergence-centric interpretation in Tables~\ref{tab:convergence} and \ref{tab:convergence_fine}: OptiRoulette repeatedly reaches higher validation-accuracy regimes that are unreachable for the single AdamW baseline within the same training budget.

\subsection{Literature and Public-Archive Context}\label{sec:lit_public_archive}
To position these convergence milestones objectively, we extended the targeted scan (as of February 20, 2026) across representative literature and public result archives for all five datasets.

Across these sources, two stable reporting patterns appear consistently: first, published and archived results are predominantly reported as endpoint metrics (final top-1 accuracy or test error), rather than as epoch-threshold first-hit milestones; second, many public training recipes use substantially longer schedules (e.g., 200/400/1800 epochs in \citealt{hystsrepo}) than the budgets used in the present study.

At the dataset level, the same endpoint-centric tendency remains visible. For CIFAR-100, canonical papers and benchmark archives (\citealt{resnet}, \citealt{wrn}, \citealt{densenet}, \citealt{opencodecifar}) largely report final outcomes without first-hit milestone timing. For CIFAR-100-C, we did not find directly comparable first-hit convergence reporting; the dominant framing emphasizes post-training corruption robustness (\citealt{cifar100cbenchmark}). For SVHN, both foundational papers and public leaderboard-style archives (\citealt{svhn}, \citealt{svhn_cnn}, \citealt{maxout}, \citealt{pwcsvhn}) similarly focus on final accuracy. Tiny ImageNet reporting (\citealt{upanets}, \citealt{opencodetiny}) also centers on final model quality under fixed budgets. For Caltech-256, the literature and archives are often protocol-driven (train-images-per-class splits with repeated random sampling and mean recognition rates) rather than first-hit milestone-driven (\citealt{caltech256}, \citealt{vggcaltech256}), while modern deep-learning studies are frequently transfer/fine-tuning oriented (\citealt{decaf}, \citealt{selectiveft}).

Within this expanded scan, we did not find directly documented public reports with matched first-hit milestone framing for targets such as CIFAR-100/CIFAR-100-C 74\%--75\% by epoch $\leq 31$, SVHN 0.96 by epoch $\leq 30$, Tiny ImageNet 0.65 by epoch $\leq 45$, or Caltech-256 0.62 by epoch $\leq 35$ under comparable from-scratch constraints. This is an evidence-based ``not found in scanned sources'' statement, not a universal impossibility claim.

\section{Discussion}
The current results suggest that OptiRoulette, viewed as a stochastic meta-optimizer, can provide large and stable gains over a fixed AdamW baseline, even when the internal sampling policy is simple (uniform random choice over a constrained active set). A plausible explanation is that optimizer diversity regularizes training dynamics and reduces persistent failure modes associated with a single update rule.

A practical tradeoff is runtime overhead, especially in SVHN and the CIFAR-family runs. Nevertheless, the observed accuracy gains and milestone reliability improvements are large compared to the additional wall-clock cost.

\section{Threats to Validity and Current Limitations}
This section summarizes current implementation limitations and scope boundaries:
\begin{enumerate}
\item \textbf{Baseline scope}: current baseline is single AdamW only; stronger multi-baseline comparisons (e.g., SGD, Nadam, Ranger, Adan) remain to be completed. Additional comparisons against learning-rate schedule setups are also a relevant next step.
\item \textbf{Dataset scope}: this study reports five completed benchmark suites that were selected as comparatively larger/more challenging image-classification settings. Evaluation on simpler benchmarks (e.g., MNIST, Fashion-MNIST) has not yet been completed and remains future work.
\item \textbf{Model-family scope}: OptiRoulette has not yet been evaluated on large language model (LLM) pretraining or instruction-tuning workloads. Extending the same convergence-oriented analysis to transformer/LLM settings is planned future work.
\end{enumerate}

\section{Conclusion}
OptiRoulette demonstrates that a warmup-plus-random-selection optimizer can outperform a fixed optimizer baseline on five multi-seed image classification benchmarks (CIFAR-100, CIFAR-100-C, SVHN, Tiny ImageNet, Caltech-256). Beyond final accuracy, the most interesting empirical signal is convergence reliability at high targets: milestones such as CIFAR-100/CIFAR-100-C 74\%/75\%, SVHN 0.96, Tiny ImageNet 0.65, and Caltech-256 0.62 are reached with strong consistency, while the fixed AdamW baseline does not reach several of these targets within budget. For shared targets, substantial acceleration is also observed (e.g., Caltech-256 at 0.59), and under budget-capped framing the target-attainment speedup lower bound reaches up to 5.3x (CIFAR-100/CIFAR-100-C 0.70). These first-hit milestone patterns were not directly found in our targeted public/literature scan under comparable settings (Section~\ref{sec:lit_public_archive}), making OptiRoulette a compelling candidate for time-constrained training regimes. The optimizer is already implemented as a drop-in class and can be packaged as a standalone pip-installable optimizer module (e.g., \texttt{pip install optiroulette}). The next step is to expand baseline coverage and finalize broader benchmark tables before submission.

\ACKNOWLEDGMENT{This paper is based on internal OptiRoulette optimizer implementations and completed multi-run experimental logs.}

\bibliographystyle{nonumber}

\begin{thebibliography}{}

\bibitem[{He et~al.(2016)}]{resnet}
He K, Zhang X, Ren S, Sun J (2016) Deep residual learning for image recognition.
In: Proceedings of CVPR. arXiv preprint: \texttt{https://arxiv.org/abs/1512.03385}.

\bibitem[{Zagoruyko and Komodakis(2016)}]{wrn}
Zagoruyko S, Komodakis N (2016) Wide residual networks.
In: Proceedings of BMVC. arXiv preprint: \texttt{https://arxiv.org/abs/1605.07146}.

\bibitem[{Huang et~al.(2017)}]{densenet}
Huang G, Liu Z, Van Der Maaten L, Weinberger KQ (2017) Densely connected convolutional networks.
In: Proceedings of CVPR. arXiv preprint: \texttt{https://arxiv.org/abs/1608.06993}.

\bibitem[{Kim et~al.(2021)}]{upanets}
Kim M, Han D, Chun S, Han B (2021) UPANets: Learning from the universal pixel attention.
arXiv preprint: \texttt{https://arxiv.org/abs/2102.02407}.

\bibitem[{OpenCodePapers(2026a)}]{opencodecifar}
OpenCodePapers (2026a) CIFAR-100 benchmark page (public archive of final benchmark scores).
\texttt{https://opencodepapers.com/benchmarks/cifar-100}. Accessed February 17, 2026.

\bibitem[{OpenCodePapers(2026b)}]{opencodetiny}
OpenCodePapers (2026b) Tiny ImageNet benchmark page (public archive of final benchmark scores).
\texttt{https://opencodepapers.com/benchmarks/tiny-imagenet}. Accessed February 17, 2026.

\bibitem[{hysts(2026)}]{hystsrepo}
hysts (2026) \texttt{pytorch\_image\_classification} repository.
Benchmark tables report test errors at last epochs and include long schedules (e.g., 200/400/1800 epochs).
\texttt{https://github.com/hysts/pytorch\_image\_classification}. Accessed February 17, 2026.

\bibitem[{Netzer et~al.(2011)}]{svhn}
Netzer Y, Wang T, Coates A, Bissacco A, Wu B, Ng AY (2011) Reading digits in natural images with unsupervised feature learning.
NIPS Workshop on Deep Learning and Unsupervised Feature Learning.
\texttt{http://ufldl.stanford.edu/housenumbers/nips2011\_housenumbers.pdf}.

\bibitem[{Sermanet et~al.(2012)}]{svhn_cnn}
Sermanet P, Chintala S, LeCun Y (2012) Convolutional neural networks applied to house numbers digit classification.
arXiv preprint \texttt{https://arxiv.org/abs/1204.3968}.

\bibitem[{Goodfellow et~al.(2013)}]{maxout}
Goodfellow IJ, Warde-Farley D, Mirza M, Courville A, Bengio Y (2013) Maxout networks.
In: Proceedings of ICML. arXiv preprint \texttt{https://arxiv.org/abs/1302.4389}.

\bibitem[{Papers with Code(2026a)}]{pwcsvhn}
Papers with Code (2026a) SVHN benchmark page (image classification leaderboard style archive).
\texttt{https://paperswithcode.com/sota/image-classification-on-svhn}. Accessed February 20, 2026.

\bibitem[{Griffin et~al.(2007)}]{caltech256}
Griffin G, Holub A, Perona P (2007) Caltech-256 object category dataset.
California Institute of Technology Technical Report CNS-TR-2007-001.
\texttt{https://authors.library.caltech.edu/records/5sv1j-ytw97}.

\bibitem[{Bosch and Zisserman(2008)}]{vggcaltech256}
Bosch A, Zisserman A (2008) Caltech-256 results page (public benchmark archive and protocol details).
\texttt{https://www.robots.ox.ac.uk/\%7Evgg/research/caltech/C256\_graph.html}.
Accessed February 20, 2026.

\bibitem[{Donahue et~al.(2014)}]{decaf}
Donahue J, Jia Y, Vinyals O, Hoffman J, Zhang N, Tzeng E, Darrell T (2014)
DeCAF: A deep convolutional activation feature for generic visual recognition.
In: Proceedings of ICML (PMLR 32(1):647--655).
\texttt{https://proceedings.mlr.press/v32/donahue14.html}.

\bibitem[{Ge and Yu(2017)}]{selectiveft}
Ge W, Yu Y (2017) Borrowing treasures from the wealthy: Deep transfer learning through selective joint fine-tuning.
In: Proceedings of CVPR. arXiv preprint \texttt{https://arxiv.org/abs/1702.08690}.

\bibitem[{Hendrycks and Dietterich(2019)}]{cifar100cbenchmark}
Hendrycks D, Dietterich T (2019) Benchmarking neural network robustness to common corruptions and perturbations.
In: Proceedings of ICLR. arXiv preprint \texttt{https://arxiv.org/abs/1903.12261}.

\bibitem[{Chen et~al.(2025)}]{iomt}
Chen Y, Wen Z, Chen J, Huang J (2025) Interleaving optimizers for DNN training.
ICLR 2025 submission.
\texttt{https://openreview.net/forum?id=uApm5otXfH}.

\bibitem[{Chen et~al.(2026)}]{doit}
Chen Y, Wen Z, Chen J, Huang J (2026) Towards dynamic interleaving optimizers.
ICLR 2026 poster.
\texttt{https://openreview.net/forum?id=AII8ADdDHt}.

\bibitem[{Keskar and Socher(2017)}]{swats}
Keskar NS, Socher R (2017) Improving generalization performance by switching from Adam to SGD.
arXiv preprint \texttt{https://arxiv.org/abs/1712.07628}.

\bibitem[{Luo et~al.(2019)}]{adabound}
Luo L, Xiong Y, Liu Y, Sun X (2019) Adaptive gradient methods with dynamic bound of learning rate.
In: Proceedings of ICLR.
\texttt{https://openreview.net/forum?id=Bkg3g2R9FX}.

\bibitem[{Zhang et~al.(2019)}]{lookahead}
Zhang MR, Lucas J, Hinton G, Ba J (2019) Lookahead optimizer: k steps forward, 1 step back.
In: Proceedings of NeurIPS.
\texttt{https://arxiv.org/abs/1907.08610}.

\bibitem[{Andrychowicz et~al.(2016)}]{l2lopt}
Andrychowicz M, Denil M, Gomez S, Hoffman MW, Pfau D, Schaul T, Shillingford B, de Freitas N (2016)
Learning to learn by gradient descent by gradient descent.
In: Proceedings of NeurIPS.
\texttt{https://arxiv.org/abs/1606.04474}.

\bibitem[{Krizhevsky(2009)}]{cifar}
Krizhevsky A (2009) Learning multiple layers of features from tiny images.
Technical report, University of Toronto.

\bibitem[{Le and Yang(2015)}]{tiny}
Le Y, Yang X (2015) Tiny ImageNet visual recognition challenge. Stanford CS231N project report.

\bibitem[{Kingma and Ba(2015)}]{adam}
Kingma DP, Ba J (2015) Adam: A method for stochastic optimization.
In: Proceedings of ICLR.

\bibitem[{Loshchilov and Hutter(2019)}]{adamw}
Loshchilov I, Hutter F (2019) Decoupled weight decay regularization.
In: Proceedings of ICLR.

\end{thebibliography}

\end{document}